\documentclass[conference]{IEEEtran}
\IEEEoverridecommandlockouts

\usepackage{cite}
\usepackage{amsmath,amssymb,amsfonts}
\usepackage{algorithmic}
\usepackage{graphicx}
\usepackage{textcomp}
\usepackage{xcolor}

\begin{document}

% Title and Authors
\title{Ultralight Signal Classification Model for Automatic Modulation Recognition\\
    \thanks{We thank the Centro para el Desarrollo Tecnológico e Industria (CDTI) for their support under the Project INSPIRE via Misiones PERTE Chip 2023. Grant EXP-00162621 / MIG-20231016.}
}

\author{
\IEEEauthorblockN{Alessandro Daniele Genuardi Oquendo}
\IEEEauthorblockA{\textit{Multiverse Computing}\\
Paseo de Miramón 170, 20014 San Sebastián, Spain}
\and
\IEEEauthorblockN{Agustín Matías Galante Cerviño}
\IEEEauthorblockA{\textit{DAS Photonics}\\
8F, Camino de Vera, Algirós, 46022 Valencia, Spain}
\and
\IEEEauthorblockN{Nilotpal Kanti Sinha}
\IEEEauthorblockA{\textit{Multiverse Computing}\\
Paseo de Miramón 170, 20014 San Sebastián, Spain}
\and
\IEEEauthorblockN{Luc Andrea}
\IEEEauthorblockA{\textit{Multiverse Computing}\\
Rue de la Croix Martre, 91120 Palaiseau, France}
\and
\IEEEauthorblockN{Román Orús}
\IEEEauthorblockA{\textit{Multiverse Computing}\\
Paseo de Miramón 170, 20014 San Sebastián, Spain \\
\textit{DIPC (Donostia International Physics Center)}\\
Paseo Manuel de Lardizabal 4, 20018 San Sebastián, Spain \\
\textit{Ikerbasque Foundation for Science}\\
Maria Diaz de Haro 3, 48013 Bilbao, Spain}
\and
\IEEEauthorblockN{Sam Mugel}
\IEEEauthorblockA{\textit{Multiverse Computing}\\
192 Spadina Avenue, 509 Toronto, Canada}
}

\maketitle

% Abstract Section
\begin{abstract}
\noindent
\textit{The growing complexity of radar signals demands responsive and accurate detection systems that can operate efficiently on resource-constrained edge devices. Existing models, while effective, often rely on substantial computational resources and large datasets, making them impractical for edge deployment. In this work, we propose an ultralight hybrid neural network optimized for edge applications, delivering robust performance across unfavorable signal-to-noise ratios (mean accuracy of 96.3\% at 0 dB) using less than 100 samples per class, and significantly reducing computational overhead.}
\end{abstract}

% Paper Content Sections
\section{Introduction}

\subsection{Motivation}
Automatic Modulation Recognition (AMR) is a crucial capability in defense applications, particularly within the field of Signals Intelligence (SIGINT). It enables the identification and classification of modulation schemes from non-cooperative emitters, often in contested or adversarial environments. This capability is essential for maintaining situational awareness, securing communication channels, and identifying potential threats. AMR systems are indispensable for modern defense operations, allowing for real-time decision-making in scenarios where signals may be deliberately obfuscated or camouflaged.

In civilian communications, AMR plays a vital role in optimizing wireless systems, such as cognitive radio networks. These systems dynamically adapt communication parameters to enhance spectral efficiency, reduce interference, and support seamless connectivity in increasingly crowded frequency bands. As wireless devices proliferate and demands for reliable, high-speed communication grow, AMR's ability to efficiently process and classify diverse signal types is central to the functionality of modern networks.

The complexity of AMR stems from the need for robust, real-time performance in challenging environments, such as those with high noise, interference, or limited computational resources. The integration of machine learning and deep learning has significantly improved AMR's accuracy and adaptability, but challenges remain. Addressing these challenges is essential to fully realizing the potential of AMR systems in both defense and civilian contexts, ensuring secure and efficient communication in a rapidly evolving technological landscape \cite{ref1, ref2, ref3}.

\subsection{Related Works}
AMR leverages advanced algorithms and neural network architectures for feature extraction and classification, achieving high recognition accuracy and computational efficiency \cite{ref4, ref5}. The integration of deep learning (DL) techniques has significantly enhanced AMR systems, addressing the limitations of traditional modulation detection methods \cite{ref6}. DL-based approaches use deep neural networks (DNNs) to extract robust features from signal data, improving recognition accuracy and reducing false alarm rates. Recent innovations, such as lightweight models incorporating convolutional neural networks (CNNs) and gated recurrent units (GRUs), aim to balance accuracy with computational efficiency, making them more viable for real-world applications \cite{ref6, cai2024semisupervised}.

Neural networks, including CNNs and recurrent neural networks (RNNs), play a pivotal role in signal processing tasks like AMR. CNNs excel at learning spatial hierarchies of features, while RNNs capture temporal dependencies in sequential data \cite{ref4}. Hybrid architectures combining CNNs and RNNs leverage the strengths of both, providing a comprehensive understanding of signal data and enhancing AMR system performance \cite{ref13}. These advances have driven the field forward, achieving higher performance and robustness in diverse signal conditions \cite{ref4}.

\subsection{Limitations and Challenges}
A major obstacle in advancing AMR technologies is the scarcity of realistic datasets, as synthetic alternatives often fail to replicate the complexity of real-world signal environments. This limitation underscores the difficulty in bridging the gap between development and reliable deployment.

The choice between analog and digital solutions further complicates the landscape. Digital systems, while highly adaptable and scalable, are more susceptible to cyberattacks like spoofing and jamming. Analog systems, in contrast, offer resilience to such threats but struggle to meet the demands of modern, dynamic communication standards.

To mitigate these challenges, techniques such as data augmentation, transfer learning, and knowledge distillation have shown promise, enabling models to perform effectively even with limited labeled data. Addressing these limitations remains critical for unlocking the full potential of AMR systems in real-world applications \cite{ref5, ref2}.

\subsection{Contribution}
This paper introduces an ultralight hybrid neural network model (XLW-CNN-LSTM) tailored for AMR on resource-constrained edge devices. By addressing the computational challenges of existing AMR solutions, we deliver a lightweight and deployable model without compromising recognition accuracy. The key contributions of this work are:
\begin{itemize}
    \item Achieved an exceptional worst-case accuracy of 93.8\% and an average-case accuracy of 96.3\% on a limited synthetic dataset (60 labeled original samples in the training set) at 20 dB Signal-to-Noise Ratio (SNR), ensuring reliable detection even under constrained conditions.
    \item Developed a novel method for sample-specific error rate identification. This enabled precise and efficient data augmentation, significantly enhancing accuracy without increasing model size or complexity. This is, to the best of our knowledge, the first AMR model optimized for both edge deployment and photonic circuit implementation.
    \item Demonstrated the model's robustness over a wide range of SNR, from -20 dB to +20 dB, outperforming larger state-of-the-art models across the electromagnetic spectrum while maintaining minimal computational overhead.
\end{itemize}

\section{Data Specifications}
\subsection{Modulated Radar Signals}
A radar signal consists of a carrier wave and a modulated wave. The carrier wave determines the frequency at which the signal is transmitted and supports the modulated wave, which conveys the radar’s information. This modulated wave is composed of a series of pulses, structured through two types of modulation: intrapulse modulation, which pertains to variations within each pulse, and interpulse modulation, which involves the temporal intervals between consecutive pulses.

In this work, we focus exclusively on intrapulse modulation, as it encompasses a broad variety of modulating functions and holds significant diversity within radar and telecommunications applications. Interpulse modulation, while important, is excluded here due to its relative simplicity in analysis and application.

We selected 11 modulation types (classes) for our analysis: \textit{BFSK, BPSK, Frank, LFM, P1, P2, P3, P4, QFM, T1, T2}. These classes encompass a diverse range of signal modulation techniques critical for evaluating recognition performance under varying noise levels:

\begin{itemize} %[label=--,leftmargin=*]
    \item \textbf{Linear Frequency Modulation (LFM)}: Pulse frequency varies linearly, commonly used in radar for range resolution.
    \item \textbf{Binary Frequency Shift Keying (BFSK)}: Alternates between two distinct frequencies, robust against noise \cite{wolff2024b}.
    \item \textbf{Binary Phase Shift Keying (BPSK)}: Similar to BFSK but with phase-based modulation, widely used in communication systems.
    \item \textbf{Frank}: Characterized by discrete phase variations following a predefined code \cite{wolff2024c}.
    \item \textbf{P1, P2, P3, P4}: Phase modulations with varying complexity; P2 and P3 include additional phase or frequency jumps.
    \item \textbf{Quadratic Frequency Modulation (QFM)}: Frequency varies quadratically.
    \item \textbf{T1, T2}: Time-frequency modulations with distinct temporal and spectral features.
\end{itemize}

\subsection{Dataset Generation}
There are 100 labeled samples per class, each with the modulated real signal over time (2048 points). The sampling frequency can be chosen arbitrarily. In terms of noise, we have white Gaussian noise between -20 dB and 20 dB of SNR.

\begin{table*}[h!]
\centering
\caption{Instantaneous Phases of Modulation Types with Parameter Ranges Used in Dataset}
\label{tab:radar_phases}
\renewcommand{\arraystretch}{1.3}
\begin{tabular}{|c|c|c|}
\hline
\textbf{Modulation type} & \textbf{Parameter range} & \textbf{Instantaneous phase, $\varphi(t)$} \\ \hline
LFM & $f_{\text{ini}} \in [0.01, 0.45]$ & $2\pi (f_0 t \pm \frac{\mu}{2} t^2)$ \\  
    & $\text{\(\Delta f\)} \in [0.05, 0.4]$ & \\ \hline
QFM & $f_{\text{min}} \in [0.01, 0.4]$ & $2\pi f_0 t \pm \pi \mu \left(t - \frac{T}{2} \right)^3$ \\
    & $\text{$\Delta f$} \in [0.05, 0.3]$ & \\ \hline
BPSK & $f_0 \in [0.05, 0.45]$ & $2\pi f_0 t + \theta$, $\theta \in \{0, \pi\}$ \\
     & $N \in \{5, 7, 11, 13\}$ & \\ \hline
BFSK & $f_1, f_2 \in [0.05, 0.45]$ & $2\pi f t$, $f \in \{f_1, f_2\}$ \\
     & $N \in \{5, 7, 11, 13\}$ & \\ \hline
T1 & $f_{0} \in [0.05, 0.45]$ & $2\pi f_0 t + \text{mod}\left\{\frac{2\pi}{n} \text{INT} \left[(mt - jT)\frac{jn}{T}\right], 2\pi\right\}$ \\
                            & $\text{\(\Delta f\)} \in [0.05, 0.4]$ & \\ \hline
T2 & $f_{0} \in [0.05, 0.45]$ & $2\pi f_0 t + \text{mod}\left\{\frac{2\pi}{n} \text{INT} \left[(mt - jT)\left(\frac{2j-m+1}{T}\right)\frac{n}{2}\right], 2\pi\right\}$ \\
   & $\Delta f \in [0.05, 0.4]$ & \\ \hline
Frank & $f_0 \in [0.1, 0.4]$ & $2\pi f_0 t + \frac{2\pi (i-1)(j-1)}{M}$ \\
      & $M \in \{6, 7, 8\}$ & \\ \hline
P1 & $f_0 \in [0.1, 0.4]$ & $2\pi f_0 t - \pi \frac{[M - (2j-1)][(j-1)M + (i-1)]}{M}$ \\
   & $M \in \{6, 7, 8\}$ & \\ \hline
P2 & $f_0 \in [0.1, 0.4]$ & $2\pi f_0 t - \pi \frac{[2i-1-M][2j-1-M]}{2M}$ \\
   & $M \in \{6, 7, 8\}$ & \\ \hline
P3 & $f_0 \in [0.1, 0.4]$ & $2\pi f_0 t + \pi \frac{(i-1)^2}{N_c}$ \\
   & $N_c \in \{6, 7, 8\}$ & \\ \hline
P4 & $f_0 \in [0.1, 0.4]$ & $2\pi f_0 t + \pi \left[\frac{(i-1)^2}{N_c} - (i-1)\right]$ \\
   & $N_c \in \{6, 7, 8\}$ & \\ \hline

\end{tabular}
\end{table*}

\begin{figure*}[h!]
    \centering
    \includegraphics[width=0.9\textwidth]{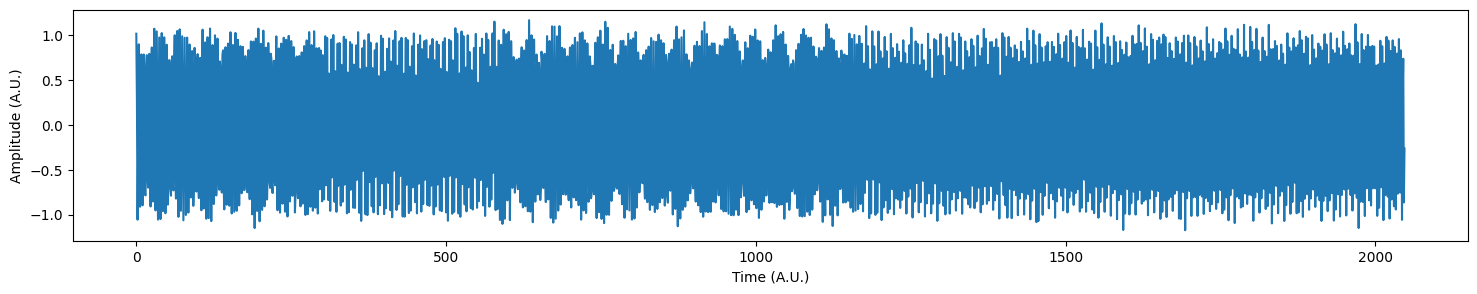}
    \makebox[0.9\textwidth]{(a) BFSK}
    
    \vspace{10pt}
    
    \includegraphics[width=0.9\textwidth]{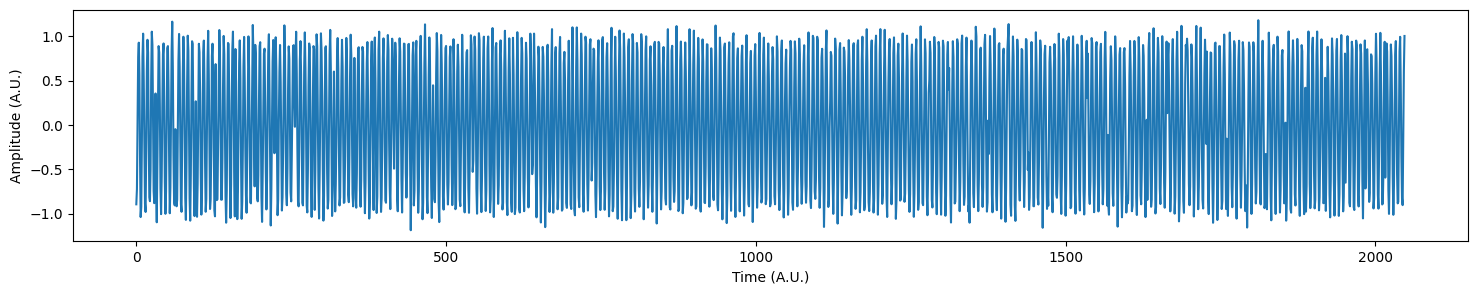}
    \makebox[0.9\textwidth]{(b) P4}
    
    \caption{Examples of intrapulse signals: (a) BFSK signal, (b) P4 signal.}
    \label{fig:intrapulse_signals}
\end{figure*}

\subsection{Data Preprocessing}
In Figure~\ref{fig:intrapulse_signals}, it is evident that distinguishing the modulation frequency directly from the raw signal is non-trivial. For instance, in BFSK, two distinct patterns—reflecting its binary nature—are observable. However, for more complex modulations such as T1 and P4, visual differentiation becomes nearly impossible, and this is true for models that operate directly on raw signals, requiring bigger architectures.

To address this challenge, the literature has explored various preprocessing techniques, including constellation diagrams \cite{1162950}, spectrograms \cite{electronics12040920}, and generalized time-frequency representations \cite{cordero2024unifiedapproachtimefrequencyrepresentations}. These methods transform the data into formats better suited for neural networks, besides enabling the application of advanced image processing techniques. 

\begin{figure*}[!t]
    \centering
    \begin{tabular}{cccc}
        \includegraphics[width=0.22\textwidth]{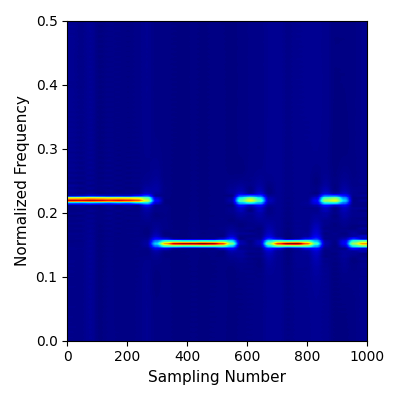} &
        \includegraphics[width=0.22\textwidth]{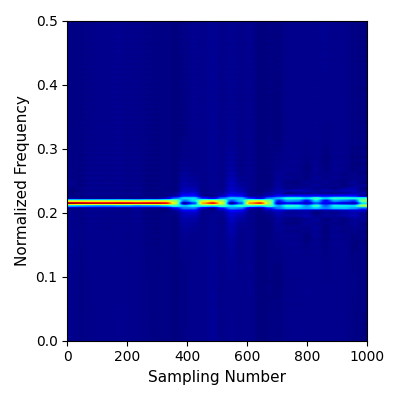} &
        \includegraphics[width=0.22\textwidth]{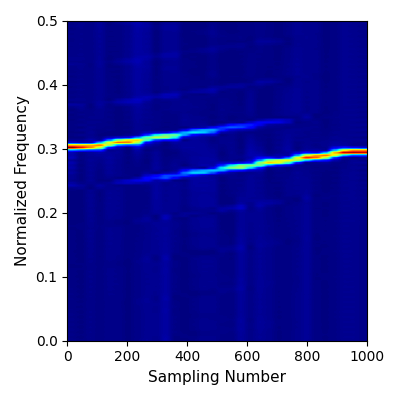} &
        \includegraphics[width=0.22\textwidth]{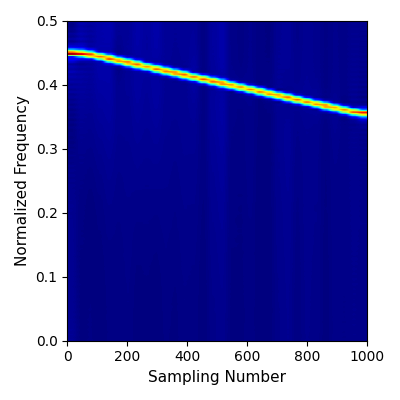} \\
        (a) BFSK & (b) BPSK & (c) Frank & (d) LFM \\
        \includegraphics[width=0.22\textwidth]{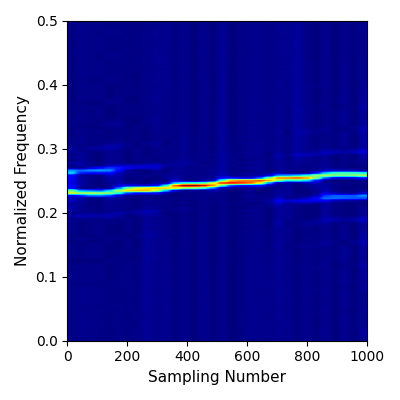} &
        \includegraphics[width=0.22\textwidth]{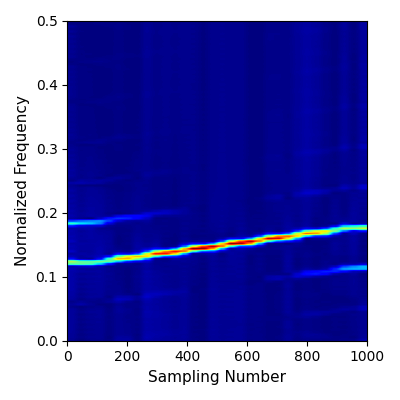} &
        \includegraphics[width=0.22\textwidth]{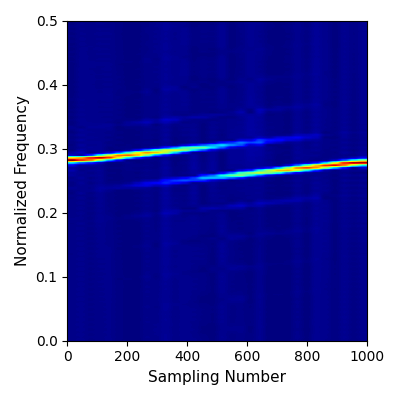} &
        \includegraphics[width=0.22\textwidth]{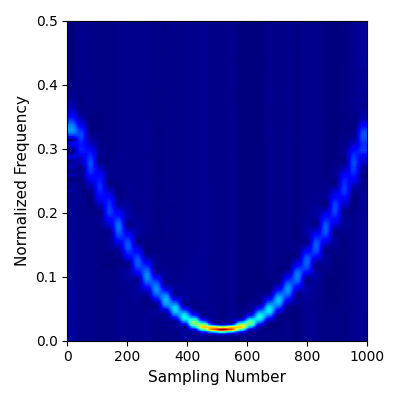} \\
        (e) P1 & (f) P2 & (g) P3 & (h) QFM \\
    \end{tabular}
    \caption{Examples of the spectrograms of intrapulse modulation signals at an SNR of 20 dB.}
    \label{fig:intrapulse_signals_grid}
\end{figure*}

\section{Methods}
\subsection{Problem Definition}
Modulation recognition can be framed as a multi-class classification challenge, where the aim is to correctly identify the modulation type of a received signal. The received signal is usually represented by the equation:

\begin{equation}
    r(t) = s(t) + n(t) = A(t)e^{j \varphi(t)} + \mathcal{N}(0, \sigma_n)
    \label{eq:signal_model}
\end{equation}

Here, $s(t) = A(t)\mathrm{e}^{\mathrm{j} \varphi(t)}$ denotes the transmitted signal, where $A(t)$ is its instantaneous amplitude, and $\varphi(t)$ its instantaneous phase. $n(t) = \mathcal{N}(0, \sigma_n)$ represents the additive noise of the channel, which in our case is additive white Gaussian noise (AWGN) with zero mean. This noise models Johnson-Nyquist or thermal noise, which originates from random movement of electrons due to the non-zero temperature of the electronic components of the receiver device.

The modulation of the signal refers to the way its amplitude, frequency and phase change with time, either only one of them or more than one at the same time. \(A(t)\) is where amplitude modulation is expressed, and \(\varphi(t)\) where we have frequency and phase modulation. The primary objective in modulation recognition is to determine the most probable modulation class $M_i$ for the given signal $r(t)$, ensuring that the probability $\Pr\{s(t) \in M_i | r(t)\}$ is maximized. Here, each $M_i$ corresponds to a different modulation type. For ease of computation, the signal $r(t)$ is typically decomposed into its in-phase and quadrature components (I/Q).

\subsection{Time-Frequency Transforms}
The raw signal, initially in the time domain, is oftentimes transformed in some way in order to more clearly ascertain its nature. Taking it from the time domain to the frequency domain through the use of the Fourier transform is a standard way of doing this, which allows for clear distinction of, for instance, frequency values that FSK modulations take, and to acquire the signal's frequency offset, as each data point of the frequency domain signal represents the properties of a sinusoid with a specifi c frequency. However, our chosen modulations' spectral content may vary significantly with time, something which is not inferred easily in the frequency domain. Thus, time-frequency transforms were developed, which allow for an explicit representation of both temporal and frequential features of the signal, at the cost of reduced resolution in both time and frequency.

A widely adopted approach to time-frequency analysis involves generating spectrograms using methods such as the Short-Time Fourier Transform (STFT), Choi-Williams distribution (CWD), and Smoothed Pseudo-Wigner-Ville Distribution (SPWVD) \cite{ref5}. These time-frequency representations (TFR), also called time-frequency distributions or transforms, take the time domain signal to the time-frequency domain, transforming it such that it now depends on both time and frequency at once, rather than just time or frequency. The end result of these techniques is broadly the same, transforming raw signals into visually distinguishable patterns, effectively reframing the time-series classification challenge as an image recognition task suitable for CNNs. However, they each deal with noise rather differently, especially linear TFRs such as the STFT and bilinear ones like Cohen's class of TFRs, which actually generalize the Wigner-Ville distribution (WVD); the latter are generally able to better separate the signal from the noise.

In our study, we adopt Cohen's class of TFRs as our preprocessing of choice for raw signals. This transformation facilitates signal analysis and leverages advanced image processing techniques to enhance model performance (Figure~\ref{fig:intrapulse_signals_grid}). A Cohen's class of TFR is defined, for continuous signals, as:

\begin{equation}
    C(t, \omega) = \frac{1}{4\pi^2} \int^{\infty}_{-\infty} \int^{\infty}_{-\infty} AF(\tau, \nu) \phi(\tau, \nu) e^{-j\nu t - j\omega \tau} \, d\nu \, d\tau
    \label{eq:cohen_class}
\end{equation}

where \(AF(\tau, \nu)\) is the ambiguity function, defined as:

\begin{equation}
    AF(\tau, \nu) \equiv \int^{\infty}_{-\infty} r\left(u + \frac{\tau}{2}\right) r^*\left(u - \frac{\tau}{2}\right) e^{j\nu u} \, du
    \label{eq:ambiguity_function}
\end{equation}

where \(r(u + \tau/2)\) is our shifted time domain signal, like \(r^{*}(u - \tau/2)\), however the latter is also conjugated. \(\phi(\tau, \nu)\) is Cohen's kernel function, generally low-pass. This kernel function is what sets each concrete TFR apart. For example, in the case of the WVD, \(\phi(\tau, \nu) = 1\), and the CWD, \(\phi(\tau, \nu) = \mathrm{e}^{-\alpha (\eta \tau )^{2}}\).
One of the key advantages of this approach is its ability to mitigate noise through embedded preprocessing strategies, such as windowing, kernel functions, and the fine-tuning of parameters in the transformation equations (Equations~\ref{eq:cohen_class}).

\begin{figure*}[!t]
    \centering
    \begin{tabular}{cccc}
        \includegraphics[width=0.22\textwidth]{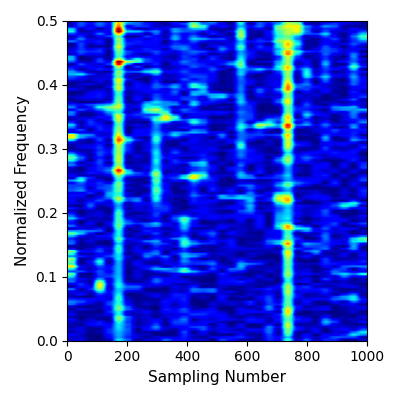} &
        \includegraphics[width=0.22\textwidth]{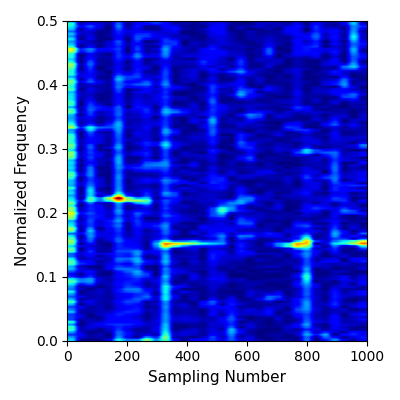} &
        \includegraphics[width=0.22\textwidth]{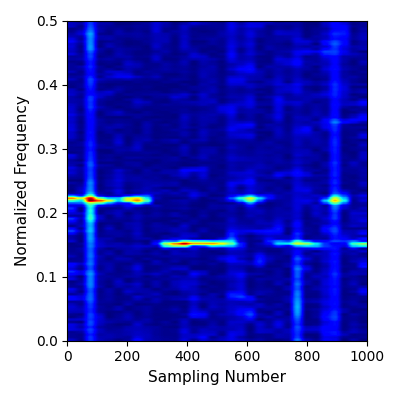} &
        \includegraphics[width=0.22\textwidth]{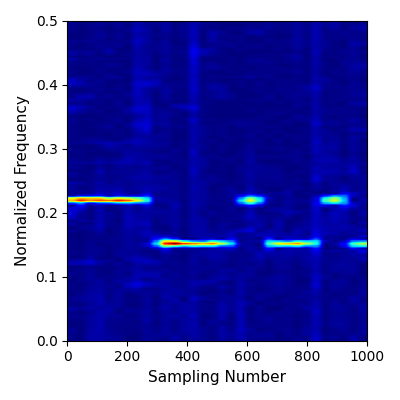} \\
        (a) BFSK, -20dB & (b) BFSK, -10dB & (c) BFSK, -5dB & (d) BFSK, 0dB \\
        \includegraphics[width=0.22\textwidth]{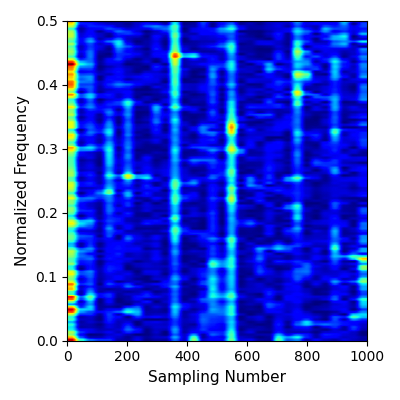} &
        \includegraphics[width=0.22\textwidth]{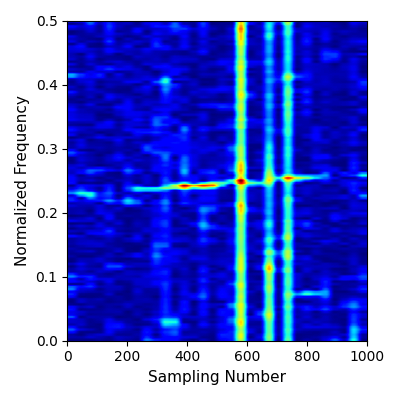} &
        \includegraphics[width=0.22\textwidth]{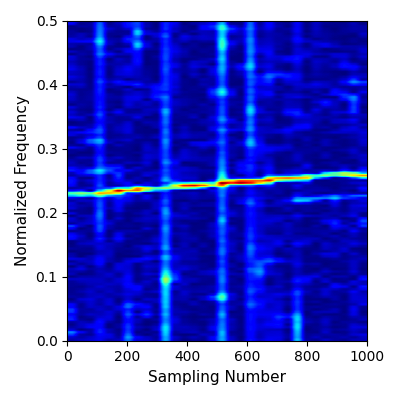} &
        \includegraphics[width=0.22\textwidth]{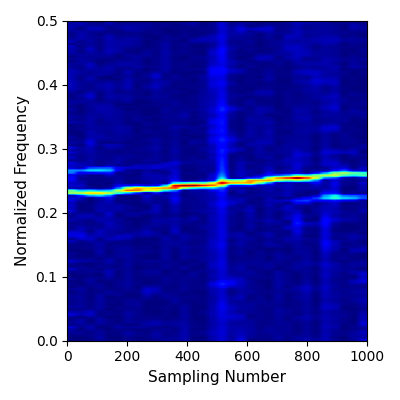} \\
        (e) P1, -20dB & (f) P1, -10dB & (g) P1, -5dB & (h) P1, 0dB \\
    \end{tabular}
    \caption{Examples of noisy modulation signals for varying SNR levels.}
    \label{fig:noisy_mods}
\end{figure*}

However, because the partitioning remains discrete, higher noise levels can introduce artificial characteristic frequencies (Figure~\ref{fig:noisy_mods}), which are not reflective of the true signal but rather numerical artifacts resulting from the computation. These artifacts can confuse the model, leading to reduced performance. For example, in cases with intense noise, the main frequency may be obscured (as seen in little presence in Figure Figure~\ref{fig:noisy_mods} (b, f) or entirely absent in Figure Figure~\ref{fig:noisy_mods} (a, e)), owing to the exponential nature of the decibel (dB) scale. Therefore, a thorough exploration of the parameter space is crucial to identifying the optimal tuning configurations that yield the best performance under various noise conditions.

\subsection{Neural Network Choice}
To address the challenges of signal classification on resource-constrained edge devices, we adopted a hybrid model that combines the strengths of 2D Convolutional Neural Networks (CNNs) and Long Short-Term Memory (LSTM) networks. CNNs efficiently extract spatial features from spectrograms but lack the ability to capture temporal dependencies, while LSTMs excel at modeling sequential relationships but add computational complexity.

The hybrid approach overcomes these limitations: the CNN extracts spatial features, and the LSTM processes them to capture temporal patterns. This design achieves high accuracy and robustness with minimal computational overhead, making it ideal for real-time deployment on edge devices.

\section{Results}

\subsection{Training}

Table \ref{tab:model_params} outlines the key parameters set during the training process, including model size, data splits, learning rate, number of training epochs, and additional optimization techniques. These settings were selected through trial and error to ensure effective learning and performance evaluation of the model.

\begin{table}[h!]
    \centering
    \begin{tabular}{|l|l|}
        \hline
        \textbf{Parameter} & \textbf{Value} \\
        \hline
        Model Parameters & 12059 (47kB) \\
        \hline
        Train-Test-Val & 0.6-0.2-0.2 \\
        \hline
        Learning Rate & 0.003 \\
        \hline
        Training Epochs & 100 \\
        \hline
        Save Best Results & True \\
        \hline
        Batch Normalization & True \\
        \hline
    \end{tabular}
    \vspace{5pt} % Add vertical space between the table and the caption
    \caption{Model Training Parameters}
    \label{tab:model_params}
\end{table}

\subsection{Outlier Detection}
An imbalance in class-specific accuracy prompted an investigation into individual outliers, a critical step in post-modelling machine learning workflows. Outlier detection is particularly useful for identifying samples that deviate from learned patterns, revealing systematic data issues or areas where the model struggles to generalize.
In our analysis, we found this process particularly effective for examining misclassified or challenging samples. Some signals were significantly disrupted and could be excluded from further consideration. Others, however, required the introduction of controlled variations through data augmentation, ensuring the preservation of their physical features while improving their detectability and representativeness in the training set.
Building on these insights, we developed and implemented a targeted data augmentation algorithm, carefully designed to enhance variability without compromising feature integrity. This approach substantially improved both global and per-class accuracy, as discussed in the next section.

% \begin{figure}[h!]
%     \centering
%     % First Image
%     \begin{minipage}[b]{\columnwidth}
%         \centering
%         \includegraphics[width=\textwidth]{figures/CNN_no_data_aug.png}
%         \\[-2pt] % Tighten space between image and caption
%         \textbf{(a)}
%     \end{minipage}
%     \vspace{10pt} % Add vertical space between the two subfigures
%     % Second Image
%     \begin{minipage}[b]{\columnwidth}
%         \centering
%         \includegraphics[width=\textwidth]{figures/CNN_yes_data_aug.png}
%         \\[-2pt]
%         \textbf{(b)}
%     \end{minipage}
%     \caption{Comparison of CNN performance: (a) Without data augmentation, (b) With data augmentation.}
%     \label{fig:cnn_data_aug_comparison}
% \end{figure}

\subsection{Model Performance}

\begin{figure}[!t]
    \centering
    \includegraphics[width=\columnwidth]{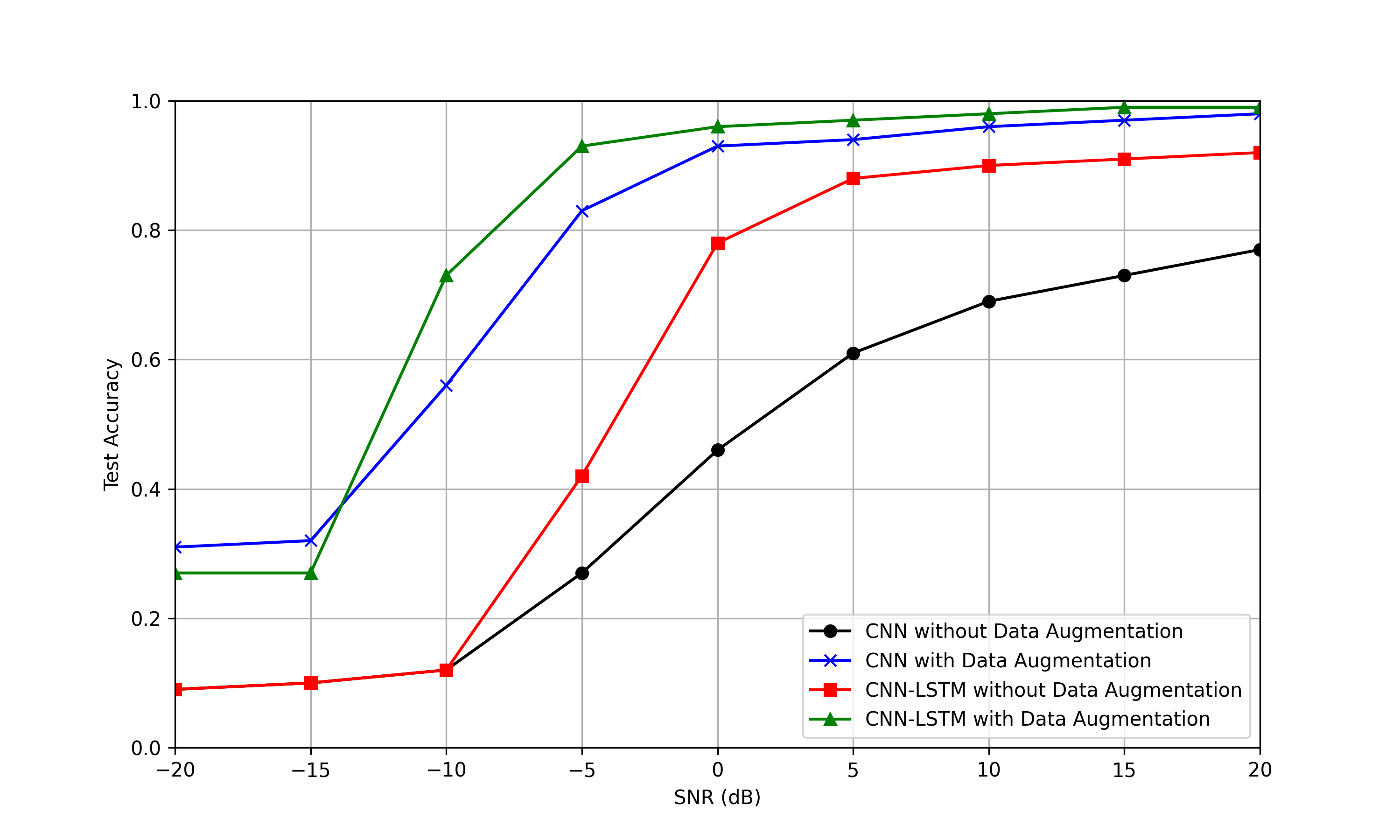}
    \caption{Performance comparison of the four model configurations.}
    \label{fig:method_comparison}
\end{figure}

When comparing the performance of the four model configurations, the combination of data augmentation and LSTM consistently outperforms the others across all noise levels. This aligns with expectations, as data augmentation increases dataset diversity and improves model generalization, the LSTM component captures more effectively temporal dependencies in the signal. Among these factors, data augmentation has a more significant impact, boosting test accuracy by an average factor of 2.43x across all noise levels, compared to a 1.26x improvement from the addition of an LSTM layer (see Fig.~\ref{fig:method_comparison}), using the bare CNN as baseline.

\begin{figure}[!t]
    \centering
    \includegraphics[width=\columnwidth]{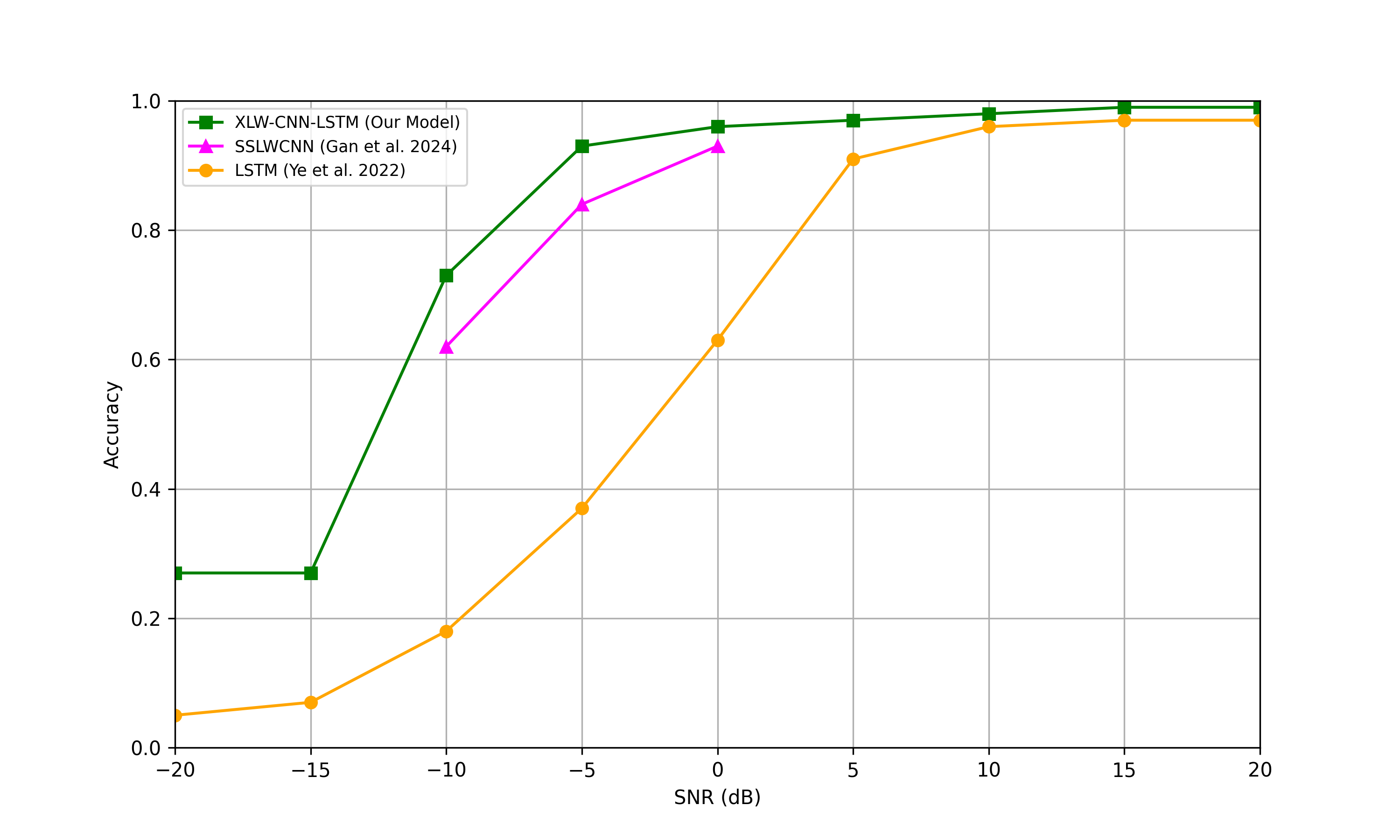}
    \caption{Comparison of our model against state-of-the-art methods.}
    \label{fig:SOTA_comparison}
\end{figure}

\begin{figure*}[htbp]
    \centering
    % First image
    \begin{minipage}[b]{0.32\textwidth}
        \centering
        \includegraphics[width=\textwidth]{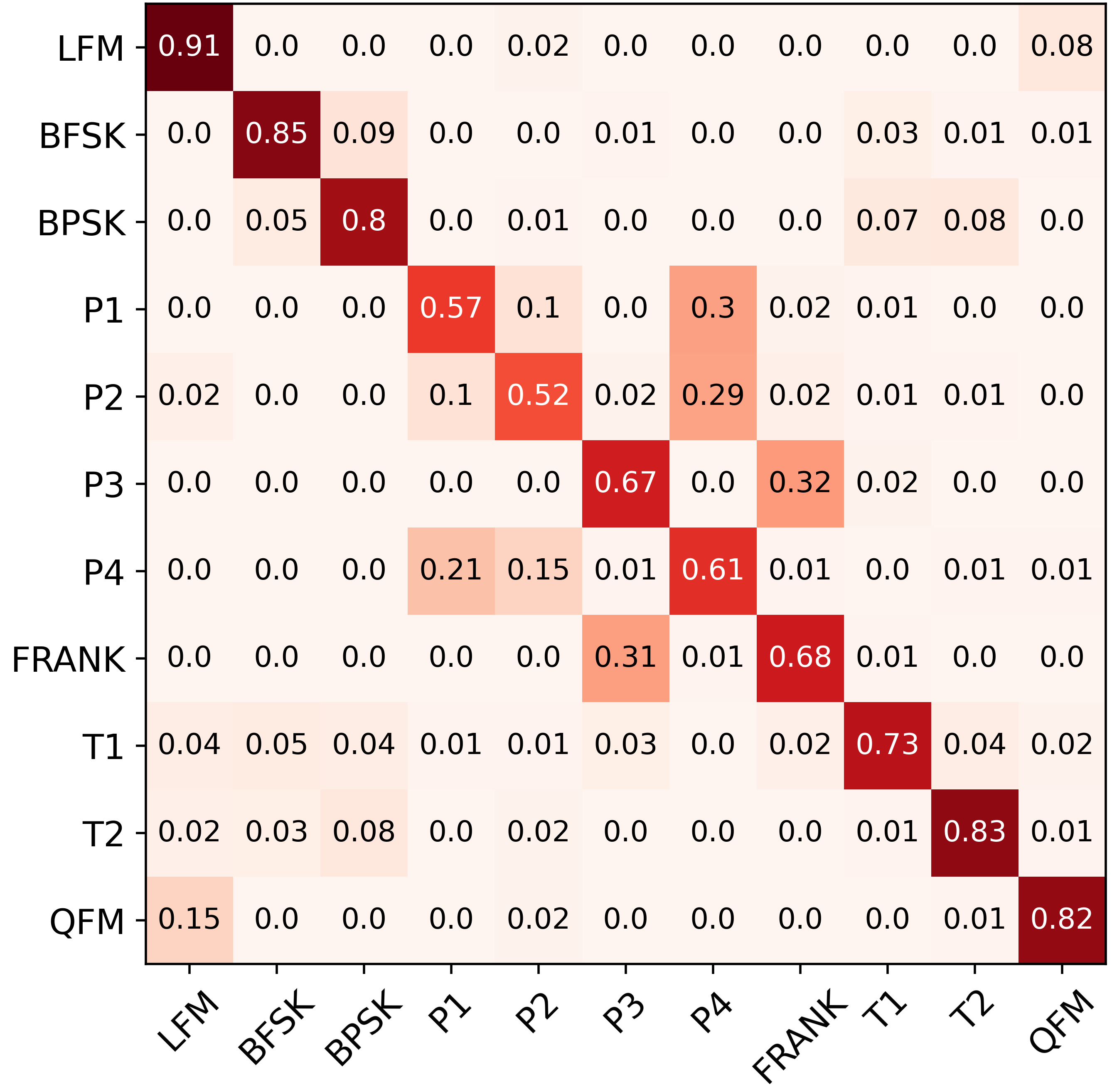}
        \\[-2pt] % Tighten space between image and caption
        \textbf{(a)}
    \end{minipage}
    \hspace{0.005\textwidth} % Minimal space between columns
    % Second image
    \begin{minipage}[b]{0.32\textwidth}
        \centering
        \includegraphics[width=\textwidth]{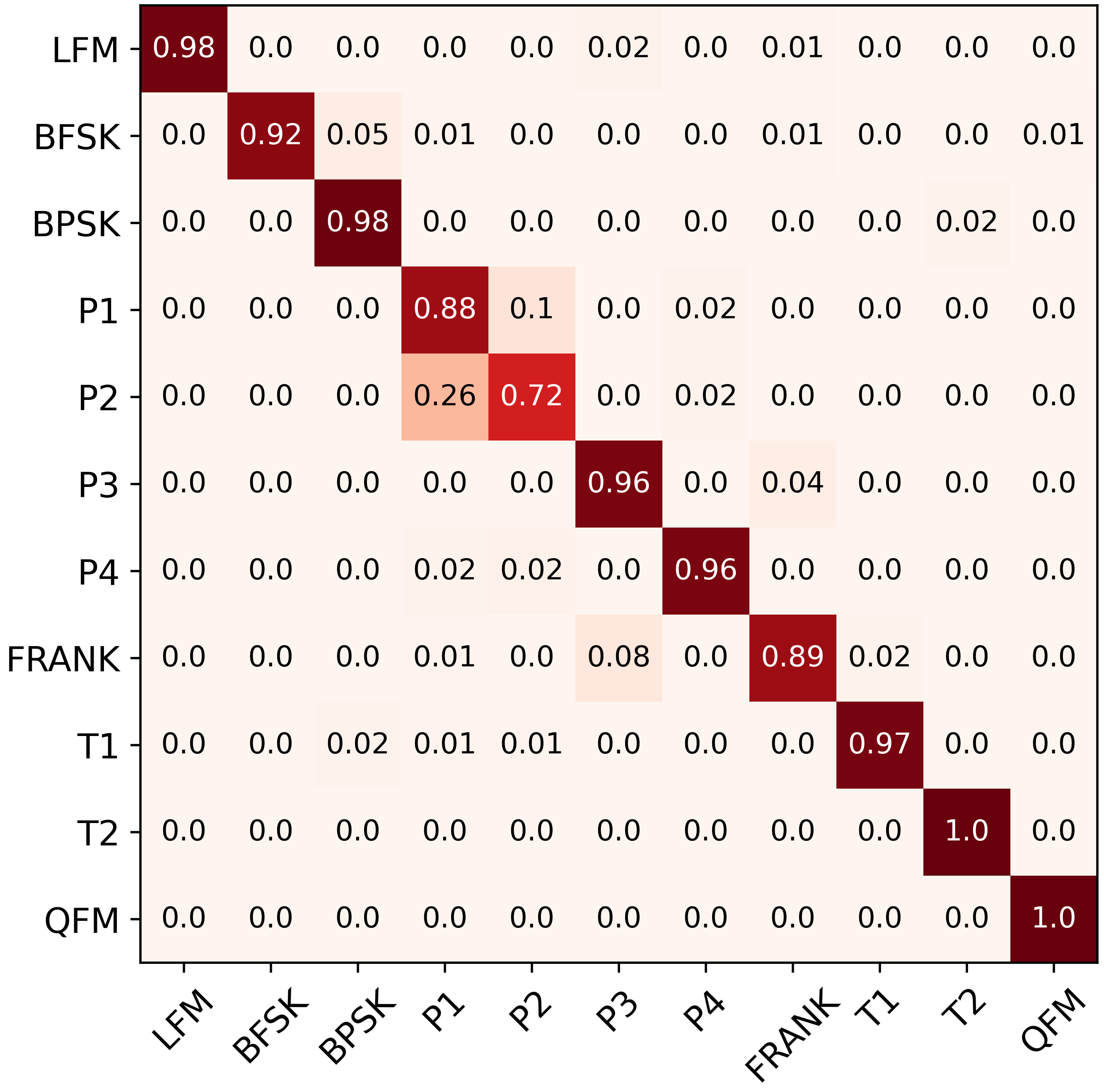}
        \\[-2pt]
        \textbf{(b)}
    \end{minipage}
    \hspace{0.005\textwidth} % Minimal space between columns
    % Third image
    \begin{minipage}[b]{0.32\textwidth}
        \centering
        \includegraphics[width=\textwidth]{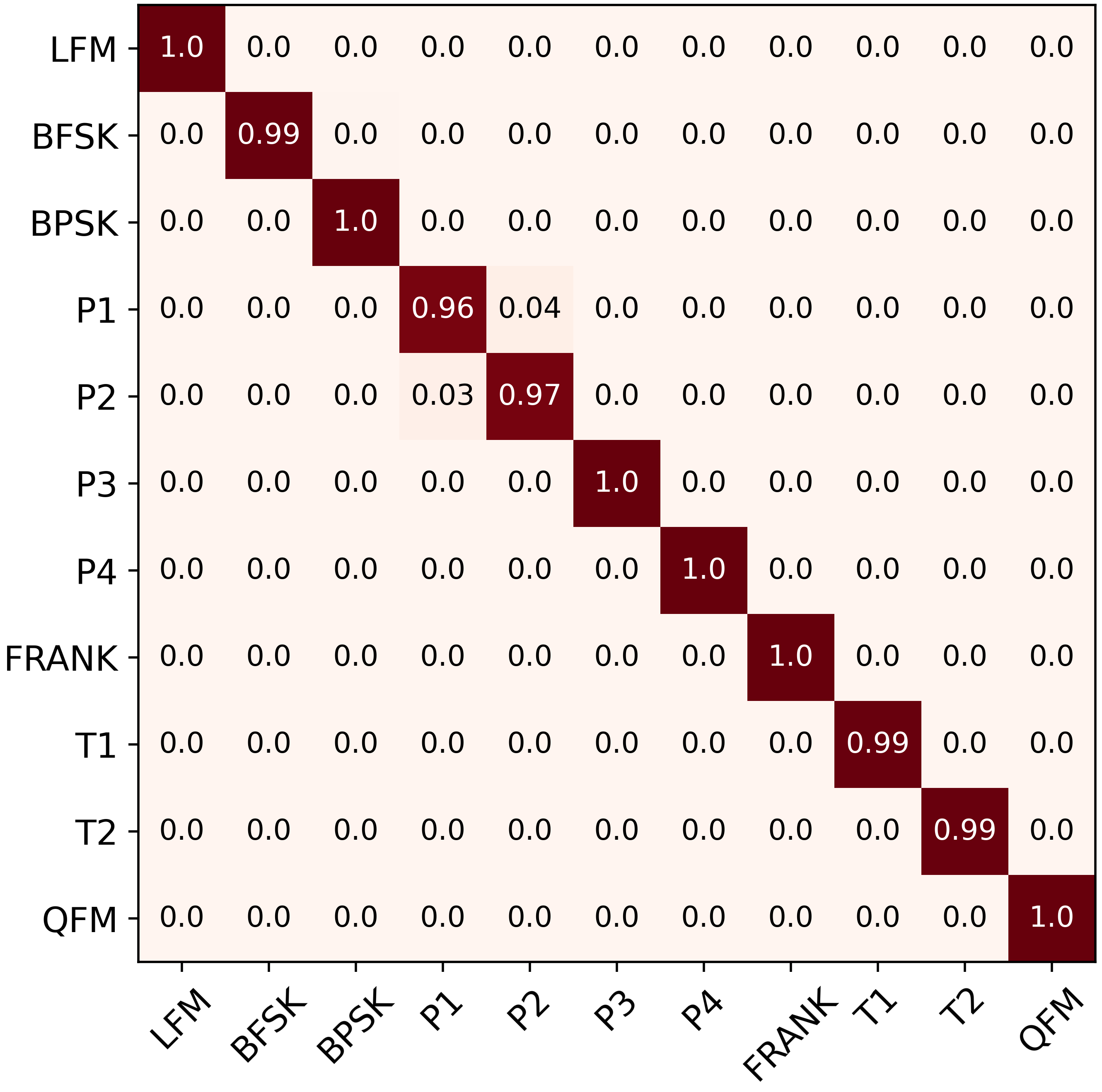}
        \\[-2pt]
        \textbf{(c)}
    \end{minipage}
    \hspace{0.005\textwidth} % Minimal space between columns
    
    \caption{Confusion Matrices for Noise Levels: -10 dB (a), -5 dB (b), and 0 dB (c).}
    \label{fig:confusion_matrices_row}
\end{figure*}

\subsection{Literature Comparison}

To surpass the state-of-the-art Gan et al. SSLWCNN model under challenging conditions—such as negative SNR and limited datasets with the same labeled sample count and type of signals—both data augmentation and LSTMs are essential. This combined approach achieves an average accuracy improvement of 1.11x over the SSLWCNN model and 2.55x over our baseline model. Notably, it achieves these results with only 1\% of the total parameters, without relying on additive techniques such as VAT (virtual adversarial training), specialized modules like RepGhost, or semi-supervised learning.
We attribute the significant improvement to the use of our bilinear Cohen’s class transform, the prioritization of data augmentation over semi-supervision, and the integration of LSTMs instead of RepGhost. While our model falls short compared to more complex architectures, such as the one employed in a Self-Attention Bidirectional-LSTM framework \cite{bilstm}, the focus of this work is to strike a balance by achieving high accuracy with a minimal number of parameters, enabling efficient deployment on edge devices (see Fig.~\ref{fig:SOTA_comparison}).

\begin{table}[h!]
    \centering
    \caption{Model Performance Comparison at 0 dB}
    \label{tab:model_comparison}
    \begin{tabular}{|l|c|c|c|l|}
        \hline
        \textbf{Model}             & \textbf{Accuracy} & \textbf{Params} & \textbf{Train Time (per epoch)} \\ \hline
        LSTM \cite{ref5}                & 0.63              & 203k           & 497s                              \\ \hline
        SSLWCNN \cite{cai2024semisupervised}             & 0.93              & 1.28M          & 256s                               \\ \hline
        \textbf{XLW-CNN-LSTM}   & \textbf{0.96}              & \textbf{12k}            & \textbf{3s}                                  \\ \hline
    \end{tabular}
\end{table}

The class-wise performance of the model is illustrated in the confusion matrices (Figure~\ref{fig:confusion_matrices_row}). When the signal strength is significantly higher than the noise, the accuracy exceeds 95\% in most cases. Even when the signal strength is comparable to the noise (Figure~\ref{fig:confusion_matrices_row}(c)), the performance remains robust. However, the behavior becomes particularly revealing when the noise dominates the signal.

At -5 dB (Figure~\ref{fig:confusion_matrices_row}(b)), the model frequently confuses P1 and P2 signals. This issue becomes more pronounced at -10 dB (Figure~\ref{fig:confusion_matrices_row}(a)), where the accuracy for these classes drops to just above 50\%. This suggests that phase-based signals, such as P1 and P2, along with Frank-coded signals, are the most challenging to distinguish under high noise levels. These challenges likely arise because these signals vary slowly and exhibit secondary frequencies that diminish as noise increases, reducing the model’s ability to differentiate them.

Despite these challenges, the results remain remarkable given the high noise levels and the lightweight nature of the model. The performance achieved surpasses that of larger architectures and more complex methods \cite{cai2024semisupervised}, highlighting the effectiveness of the proposed approach.

\begin{figure}[h!]
    \centering
    \includegraphics[width=\columnwidth]{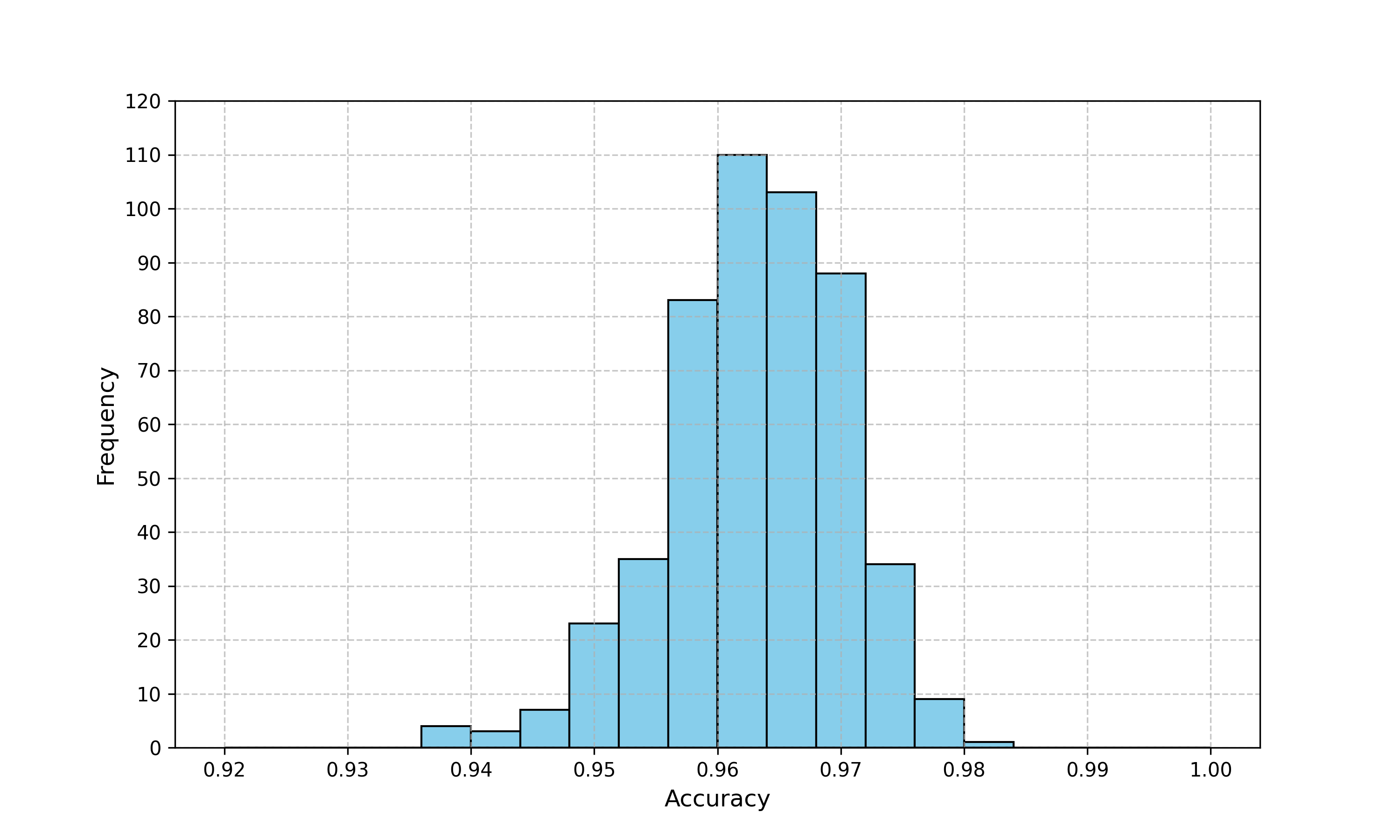}
    \caption{Histogram of model accuracy across 500 simulation runs at 0 dB SNR.}
    \label{fig:cnn_aug_histo}
\end{figure}

The results of the variance simulation for 500 runs (different test-train dataset folds) at 0 dB are shown in Figure~\ref{fig:cnn_aug_histo}. The model achieves consistent performance across all iterations, with no case falling below 93.8\% accuracy and with a 96.3\% mean accuracy. This highlights the robustness and reliability of the proposed approach, even on edge cases.

\section{Applications}
\subsection{Edge Devices}
The proposed model demonstrates significant potential for deployment on edge devices, particularly in defense applications. Its lightweight architecture allows for real-time detection and decision-making directly on-site, reducing latency and enabling rapid responses to critical situations. Additionally, deploying such models on the edge minimizes vulnerabilities associated with centralized data transmission, enhancing security and reliability. These features make it particularly suited for applications in aircraft and other resource-constrained platforms.

\begin{figure}[h!]
    \centering
    \includegraphics[width=0.9\columnwidth]{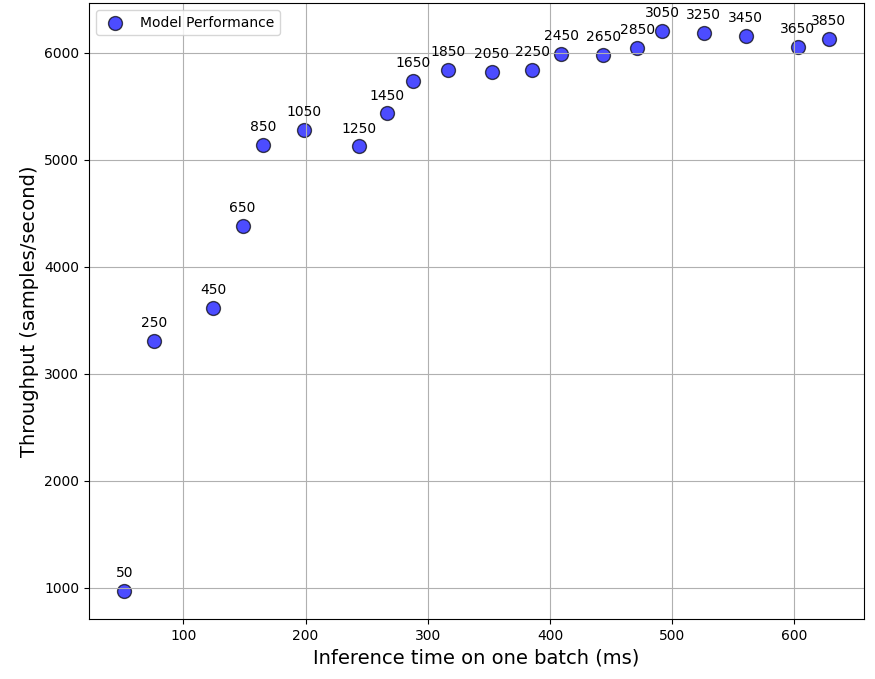}
    \caption{Latency vs. Throughput for different batch sizes. Each point is annotated with its corresponding batch size.}
    \label{fig:lat_vs_thru}
\end{figure}

The latency vs. throughput evaluation was conducted on an Apple M1 Pro chip (10-core CPU with 8 performance and 2 efficiency cores, 16GB unified memory). Despite the absence of GPU acceleration, the throughput plateaued at a reasonable level, demonstrating that the system is efficient and well-suited for edge computations even with increasing batch sizes. This result highlights the feasibility of deploying such models on edge devices where computational resources are constrained.

\subsection{Photonic Implementation}
The model’s neuron-based structure, which avoids computationally intensive squaring functions, makes it well-suited for mapping onto photonic circuits. By leveraging the inherent parallelism of photonic systems, the proposed model could achieve even faster performance, aligning with the requirements of modern, high-throughput applications. \cite{peserico2023}

Exploring such a photonic implementation in future work could pave the way for photonic computing solutions, bringing together speed, efficiency, and scalability in resource-constrained environments \cite{feldmann2019, feldmann2021}. 

\section{Conclusion}
In this work, we presented a highly efficient and lightweight hybrid neural network for Automatic Modulation Recognition (AMR), achieving a remarkable worst-case accuracy of 93.8\% and an average-case accuracy of 96.3\% at 0 dB SNR on a limited and balanced dataset. This performance is especially critical for defense applications, where even a single misclassification could have catastrophic consequences.

Through sample-specific error analysis, we identified and implemented targeted data augmentation strategies that significantly improved accuracy without increasing the model's complexity or size. This approach not only reduced the total number of parameters but also made the model well-suited for deployment on resource-constrained edge devices, including photonic circuits, where computational resources are limited.

The proposed model demonstrated robustness across the electromagnetic spectrum, outperforming state-of-the-art models while maintaining a fraction of their size and complexity. This achievement paves the way for practical real-time applications in both defense and civilian scenarios. With further exploration of custom kernel transforms, advanced noise filtering techniques, and targeted augmentation methods, we can expect to enhance the model's performance, especially at challenging SNR levels such as -10 dB.

Future work will focus on implementing the proposed architecture on actual microprocessors or photonic chips to validate its concept, unlocking new possibilities for fast, efficient, and reliable signal classification on edge devices.

% \newpage
% References
\bibliographystyle{IEEEtran}
\bibliography{bibliography}

\end{document}